\documentclass[letterpaper, 10 pt, conference]{ieeeconf}  
\IEEEoverridecommandlockouts                              
\usepackage{url}
\usepackage{amsmath,amssymb}
\usepackage{graphicx,color}
\usepackage{verbatim}
\usepackage{gensymb}
\usepackage{subfig}
\usepackage{siunitx}
\usepackage{adjustbox}
\usepackage{algorithm} 
\usepackage{algorithmicx, algpseudocode} 
\graphicspath{{figures/}}

\usepackage{caption}
\def\tablename{Table}

\title{\LARGE \bf
Applying Depth-Sensing to Automated Surgical Manipulation \\ with a da Vinci Robot
}

\author{Minho Hwang$^{*,1}$, Daniel Seita$^{*,1}$,  Brijen Thananjeyan$^1$,  Jeffrey Ichnowski$^1$,\\  Samuel Paradis$^1$,  Danyal Fer$^2$,  Thomas Low$^3$,  Ken Goldberg$^1$
\thanks{*Equal contribution.}
\thanks{$^{1}$AUTOLAB at the University of California, Berkeley, USA. {\tt \small http://autolab.berkeley.edu/}}%
\thanks{$^{2}$UC San Francisco East Bay, USA.}%
\thanks{$^{3}$SRI International, USA.}%
\thanks{Correspondence to: Minho Hwang, {\tt\small gkgkgk1215@berkeley.edu}, and Daniel Seita {\tt\small seita@berkeley.edu}}%
}
\begin{document}

\maketitle
\thispagestyle{empty}
\pagestyle{empty}

\begin{abstract}
Recent advances in depth-sensing have significantly increased accuracy, resolution, and frame rate, as shown in the 1920x1200 resolution and 13 frames per second Zivid RGBD camera. In this study, we explore the potential of depth sensing for efficient and reliable automation of surgical subtasks. We consider a monochrome (all red) version of the peg transfer task from the Fundamentals of Laparoscopic Surgery training suite implemented with the da Vinci Research Kit (dVRK). We use calibration techniques that allow the imprecise, cable-driven da Vinci to reduce error from \SIrange{4}{5}{\milli\meter} to \SIrange{1}{2}{\milli\meter} in the task space. We report experimental results for a handover-free version of the peg transfer task, performing 20 and 5 physical episodes with single- and bilateral-arm setups, respectively. Results over 236 and 49 total block transfer attempts for the single- and bilateral-arm peg transfer cases suggest that reliability can be attained with \SI{86.9}{\percent} and \SI{78.0}{\percent} for each individual block, with respective block transfer speeds of 10.02 and 5.72 seconds. Supplementary material is available at \url{https://sites.google.com/view/peg-transfer}.
\end{abstract}

\section{Introduction}\label{sec:intro}

Robotic Surgical Assistants (RSAs)
such as Intuitive Surgical's da Vinci~\cite{dvrk2014} are regularly used in hospitals and surgical procedures through teleoperation. 
The introduction of RSAs gave surgeons greater ability to perform complex tasks through improved dexterity and visualization. This increased the number of surgeons able to offer minimally invasive surgery to patients, reducing their post operative pain and length of stay in the hospital compared to open surgery~\cite{danyal_fer_suggestion2,danyal_fer_suggestion}. RSAs also provide a platform for the automation of some surgical tasks, which have potential to aid surgeons by reducing fatigue or tedium~\cite{yip2017robot}.

We consider the well-known peg transfer task from the Fundamentals of Laparoscopic Surgery (FLS)~\cite{fls_1998} and how it could be automated at a speed and reliability level on par with a professional surgeon.  This is an extremely high bar, as human surgeons perform this task with great dexterity~\cite{fls_2004}.  In addition, commands to cable-driven RSAs yield errors in motion due to backlash, cable tension, and hysteresis~\cite{pastor2013,miyasaka2015,kalman_filter_2016,mahler2014case,Kehoe2014}.

\begin{figure}[t]
\center
\includegraphics[width=0.45\textwidth]{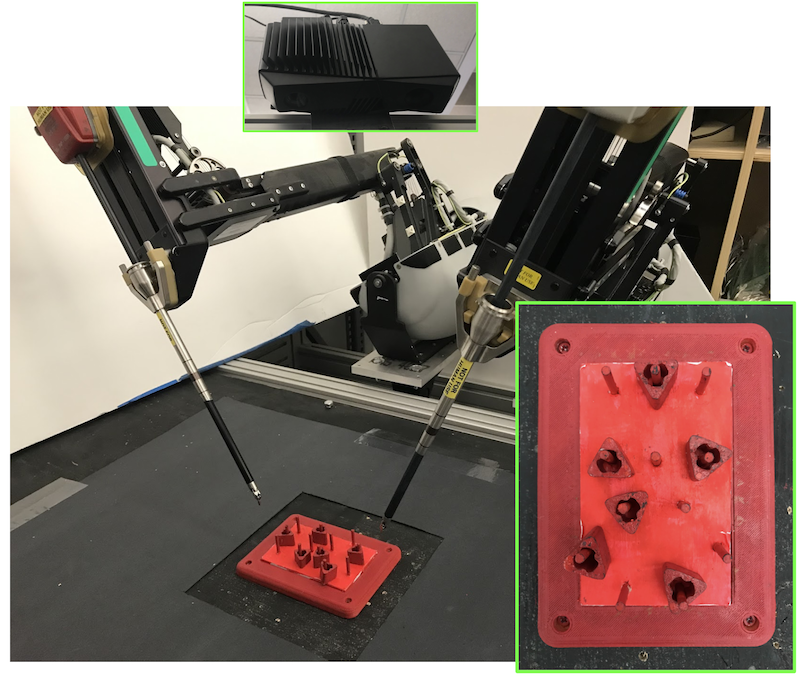}
\caption{
\small
\textbf{Automated peg-transfer task setup.} The da Vinci surgical robot utilizes either one or both of its arms to transfer blocks between pegs. The workspace (zoomed-in inset to the bottom right) consists of six hollow triangular blocks on a peg board for a procedure in progress. The task starts with all six blocks on the left side of the board. One must transfer the blocks to the right side of the board, and then bring them back to the left side. The blocks, pegs, and peg board are monochrome (painted red) to simulate a surgical setting. We use an overhead Zivid RGBD camera.
}
\vspace*{-10pt}
\label{fig:money-shot}
\end{figure}

In this task, the surgeon transfers six hollow triangular blocks from pegs on one half of a board to the pegs on the other half (see Fig.~\ref{fig:money-shot} inset).
To the best of our knowledge, the only prior work that focuses on automating a version of the peg transfer task is by Rosen and Ma~\cite{auto_peg_transfer_2015}, who used a single Raven II~\cite{raven2013} arm to perform handover-free peg transfer with three blocks. The present paper revisits this pioneering work and applies depth sensing to a monochrome variant of the peg transfer task using six blocks and the da Vinci Research Kit (dVRK)~\cite{Ballantyne2003,dvrk2014}.

To sense the blocks and pegs, we use a Zivid One Plus RGB+depth (RGBD) camera which can provide 1920x1200 pixel images at 13 frames per second with depth resolution \SI{0.5}{\milli\meter}. RGBD technology is advancing\footnote{\url{https://bair.berkeley.edu/blog/2018/10/23/depth-sensing/}} and is widely used in industrial automation. Additionally, the size of RGBD sensors is not a severe restriction in open surgical environments, and robots such as the Taurus from SRI International are being developed for this purpose~\cite{madapana2019desk}. While depth sensing is invaluable for robotics applications such as data-driven grasping~\cite{mahler2017dexnet}, we are unaware of prior applications in RSAs, as depth sensing is not yet available for minimally-invasive surgery but should be considered for open-body tele-surgical systems operated remotely via intermittent supervision, where limited autonomy may be necessary.

This paper contributes: 1) the first application of depth-sensing to an RSA, 2) a robust depth-sensing perception algorithm for peg transfer, and 3) results on 20 and 5 episodes of automated peg transfer with single and bilateral arms, respectively, suggesting reliability of \SI{86.9}{\percent} and \SI{78.0}{\percent}, with transfer speeds of 10.02 and 5.72 seconds per block. Code and videos are available at \url{https://sites.google.com/view/peg-transfer}.

\section{Related Work}\label{sec:rw}

Although surgical robotics has a long history, summarized in surveys~\cite{Moustris2011,Beasley2012,surgical_robotics_2016}, no procedures are fully autonomous due to uncertainty in perception and control.

\subsection{Autonomous Robot Surgery} 

In non-clinical research settings, several groups have explored autonomous robot surgery~\cite{yip2017robot}. Key tasks of interest, some of which are part of the FLS curriculum, include pattern cutting~\cite{murali2015learning,thananjeyan2017multilateral,intermittent_synch_2018}, suturing~\cite{sen2016automating,rosen_icra_suturing_2017,saeidi_suturing_icra_2019,Schulman2013, thananjeyan2019safety}, debridement for rigid and soft objects in static~\cite{Kehoe2014,mahler2014case,seita_icra_2018} and dynamic~\cite{surgical_camera_2018,stewart_platform} cases, needle extraction and insertion~\cite{extraction_needles_2019,needle_insertion_deformation_2019,automated_needle_pickup_2018}, knot-tying~\cite{vandenBerg2010,knot_tying_2002,improved_knots_case_2013}, and tumor localization~\cite{mckinley2015,garg2016gpas,mckinley2016}.

Automation in robot surgery has catalyzed the development of novel techniques in trajectory optimization and planning~\cite{Osa-RSS-14}, compliance manipulation~\cite{compliance_2018}, 3D reconstructions of deformable objects~\cite{super_yip_2019}, simulators or demonstrator data for imitation learning~\cite{seita_fabrics_2019}, reinforcement learning~\cite{thananjeyan2017multilateral,rosen_icra_tissues_2019,rosen_tissues_qlearn_2019}, or task segmentation~\cite{tsc-dl-2016,krishnan2017ddco,madapana2019desk,rahmantransferring}.

One challenge with autonomous minimally invasive surgery is that commercial robot systems, such as the Raven II~\cite{raven2013} and the da Vinci~\cite{dvrk2014}, are cable-driven and thus can have inaccurate motion and actuation~\cite{pastor2013,miyasaka2015,kalman_filter_2016,mahler2014case,Kehoe2014,seita_icra_2018} due to their susceptibility to cable stretch, backlash, hysteresis, and decalibration. We present a calibration procedure in Section~\ref{ssec:calibration} which was able to reliably achieve accuracy within 1--2~mm of error in the workspace.

\subsection{Peg Transfer Task}


The peg transfer task is one of the five tasks in the Fundamentals of Laparoscopic Surgery (FLS)~\cite{fls_2004}. The goal is to transfer six rubber triangular blocks from one half of a pegboard to the other, and then back. Transferring a single block requires grasping the block, lifting it off of a peg, handing it to the other arm, moving it over another peg, and then placing the block around the targeted peg. In this paper, we do not consider the handoff step because it requires large wrist motions --- the gripper that picks up the block also places it on the target peg. 
Fried~et~al.~\cite{fls_2004} performed a study in which surgical residents performed the peg transfer task, which aims to assess and develop the surgeons' depth and visual-spatial perception. Task performance was measured based on completion time, with a score penalty whenever a block fell outside the surgeon's view. The study found that superior performance in peg transfer correlates with performance during real laparoscopic surgical settings, validating the rationale for having surgeons practice peg transfer.

Prior work looks into improving human performance of the peg-transfer task.
Abiri~et~al.~\cite{haptic-feedback-surgery-2019} focus on providing better haptic feedback to compensate for the loss of force feedback when teleoperating surgical robots, and test on the peg transfer task. Brown~et~al.~\cite{rating_peg_transfer_2017} provide ways to automatically evaluate a surgeon's skill at the peg transfer task by using data from contact forces and robot arm accelerations. Rivas-Blanco~et~al.~\cite{ur5-peg-transfer-2019} apply learning from demonstrations to autonomously control camera motion during teleoperated peg transfer. Other work~\cite{laparoscopic_transfer_2014} provides additional teleoperated peg transfer benchmarks or proposes novel methods to identify stages of the peg transfer task, called surgemes, across robot platforms~\cite{madapana2019desk, rahmantransferring}. None of these prior works focus on automating peg transfer. 

Rosen and Ma~\cite{auto_peg_transfer_2015} attempted to automate a variant of the peg transfer task using a Raven II~\cite{raven2013}. They used one robot arm and three blocks per episode, and transferred in one direction. They compared performance with a human using Omni VR masters to control the arms, with each doing 20 episodes, for 60 total block transfer attempts. Results showed that the autonomous robot was able to achieve nearly the same block transfer success rate (56/60) as the human (60/60), and was twice as fast ($25\pm0.0$ vs $49\pm5.7$ seconds for each episode). We use the da Vinci with depth sensing, and transfer six blocks in both directions.

\section{Problem Statement}\label{sec:PS}

\begin{figure}[t]
\center
\includegraphics[width=0.48\textwidth]{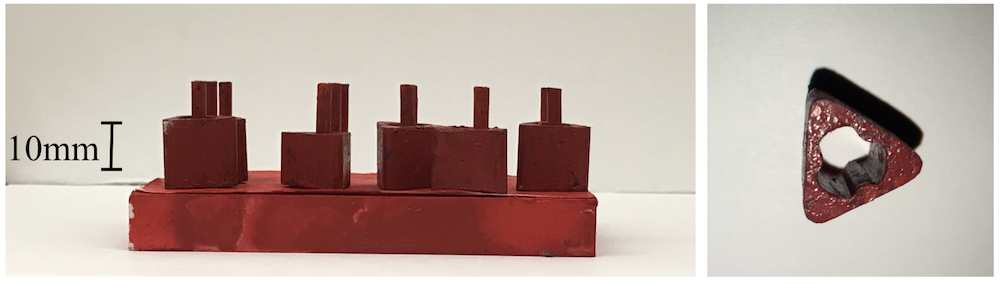}
\caption{
\small
\textbf{Block properties and sizes}. The left image shows all six blocks on a peg board, demonstrating that the height of the top layer of the blocks is not uniform. We overlay a scale of \SI{10}{\milli\meter}. The right image shows a top-down view of one of the blocks with a shadow, showing the hollow interior.
}
\vspace*{-10pt}
\label{fig:different-heights}
\end{figure}

The setup for the peg transfer task is shown in Fig.~\ref{fig:money-shot}.
The task uses six triangular rubber blocks shown in Fig.~\ref{fig:different-heights}. Each block is roughly \SI{15}{\milli\meter} in height and has triangular edges of length \SI{18}{\milli\meter} and a hollow center spanning \SIrange{5}{10}{\milli\meter}. Fig.~\ref{fig:different-heights} demonstrates that these are approximations, as the block edges are not at uniform heights.
We define one \emph{episode} of the task to be the full procedure where the surgical robot attempts to move the six blocks from the left to the right, and then moves all of them back. We test two variants: one with a single dVRK arm moving the blocks, and a bilateral one with both arms moving blocks simultaneously.

The visual cues in a surgical environment are rarely distinct and surgical decision making relies heavily on perception of minor differences in color, depth and texture. For this reason we paint the board, pegs, and blocks red to mimic real surgical settings in which tissue may be a more uniform red hue, so color may not provide a sufficient signal to automate sensing the state of the environment.

\subsection{Failure Modes}\label{ssec:failure-cases}

\begin{figure}[t]
\center
\includegraphics[width=0.48\textwidth]{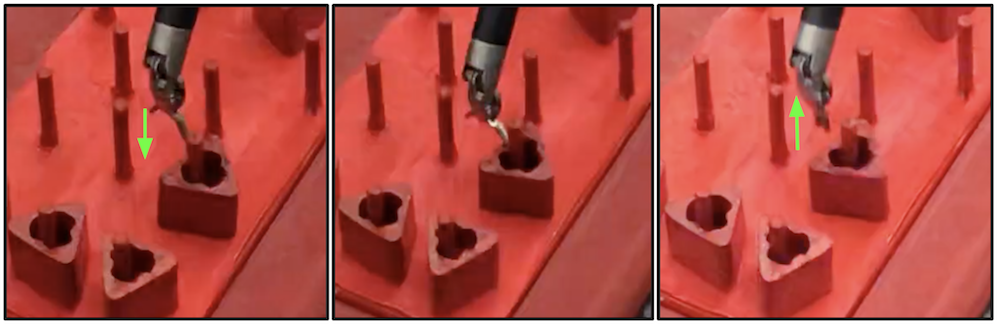}
\caption{
\small
Time lapse of an example \emph{pick} failure from the dVRK. The pick operation initially opens the gripper and lowers it (first image), but the gripper only barely touches the block edge (second image). When the gripper closes, it is unable to get the block within its grip as it rises (third image).
}
\vspace*{-6pt}
\label{fig:failure-grasp}
\end{figure}

\begin{figure}[t]
\center
\includegraphics[width=0.48\textwidth]{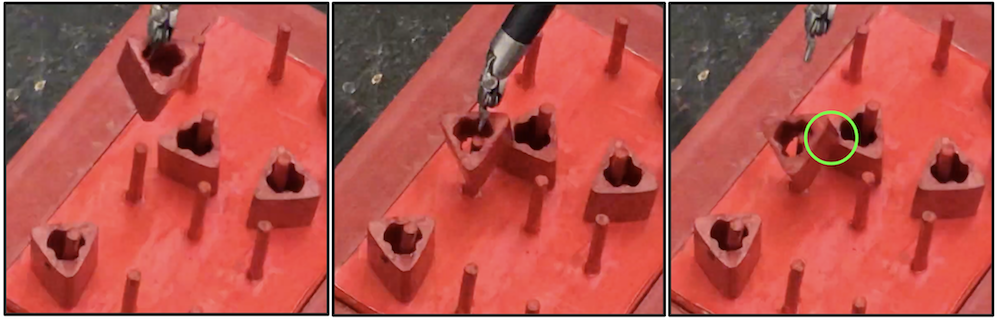}
\caption{
\small
Time lapse of an example \emph{place-stuck} failure from the dVRK. The placing operation results in the bottom of the block making contact with a block underneath (overlaid circle in third image), preventing it from being fully inserted.
}
\vspace*{-10pt}
\label{fig:failure-place-stuck}
\end{figure}

\begin{figure}[t]
\center
\includegraphics[width=0.48\textwidth]{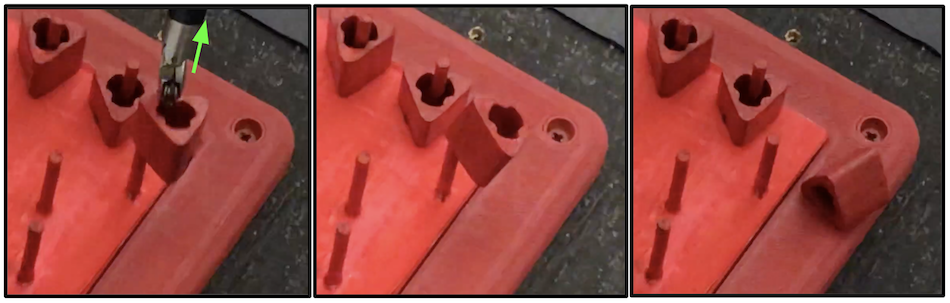}
\caption{
\small
Time lapse of an example \emph{place-fall} failure from the dVRK. The placing operation results in the bottom of the block making contact with the top of the peg. When the gripper releases the block and moves up (first images) the block eventually falls off the peg (last two images).
}
\vspace*{-10pt}
\label{fig:failure-place-fall}
\end{figure}

To better understand the performance of either the surgical robot or a human operator at the peg transfer task, we consider a set of failure modes. For consistency, we use and expand on failure definitions from prior work~\cite{auto_peg_transfer_2015}. We calculate failures based on each individual attempt at moving a block within an episode. The failure cases are: 

\begin{enumerate}
    \item \textbf{Pick failure}: an error in grasping the block from its starting peg, so that the block is not lifted free from the peg. As described in Section~\ref{ssec:error_recovery}, after this type of failure, we allow the robot one more attempt at transferring the block.
    \item \textbf{Place failure}: when a block is not fully inserted onto its target peg and does not make contact with the bottom of the workspace. We sub-divide placing failures into two categories: \textbf{place-stuck failures} for when placing results in blocks stuck on top of a peg or another block, and \textbf{place-fall failures} for when placing results in blocks that fall on the surface.
\end{enumerate}

Fig.~\ref{fig:failure-grasp} shows an example grasping failure, and Figs.~\ref{fig:failure-place-stuck} and~\ref{fig:failure-place-fall} show examples of the two placing failures, all from the automated dVRK system we use for experiments. We sub-divide placing failures because place-stuck and place-fall failures have different effects in practice. Typically it is easier to recover from place-stuck failures because the blocks still lie on top of their target peg and a gentle nudge can usually slide the block into place. In contrast, a place-fall failure requires an entirely new grasp to pick a fallen block.

\section{Robot Calibration}\label{sec:setup}

\label{ssec:calibration}

\begin{figure*}[t]
\center
\includegraphics[width=1.00\textwidth]{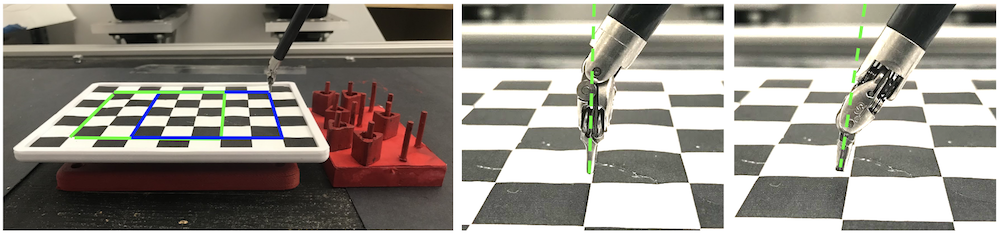}
\caption{
\small
\textbf{Calibration.} Left: when calibrating the dVRK, we use a checkerboard with each edge \SI{16}{\milli\meter} long, and move the robot's end-effector towards the corners, which are at a known height offset of the peg tips. The right arm moves to all corners included in the overlaid blue grid above, while the arm to the left would be calibrated independently and go to all points in the overlaid green grid. At a given position, however, rotating the robot's arm will result in the tip being at a different location. For example, the second image shows the tip at a checkerboard corner, and the third image shows the result when rotating the roll angle by \SI{90}{\degree}, which means the actual tip is at a spot \SIrange{2}{4}{\milli\meter} away. For this reason, we discretize the roll angle and calibrate once per roll angle discretization.
}
\vspace*{-6pt}
\label{fig:calibration}
\end{figure*}

The calibration technique we use is based on the procedure from Seita~et~al.~\cite{seita_fabrics_2019}. We calibrate the positions by using a checkerboard located at a plane that roughly mirrors the height of the blocks when they are inserted into pegs. We servo each end effector with the gripper facing down to each corner of the checkerboard and record positions. During deployment, when given an image, we map the image pixels to a 2D coordinate on the checkerboard. For a given coordinate frame, we perform bilinear interpolation to estimate the robot position from the four surrounding known points. Fig.~\ref{fig:calibration} provides an overview.

We observe that the roll angle of the end effector's pose affects the positioning of the robot (see Fig.~\ref{fig:calibration}), so we follow the method from Seita~et~al.~\cite{seita_icra_2018} and discretize the roll. We choose two discretizations of the roll value, \SI{0}{\degree} and \SI{90}{\degree}, as these are approximately close to the roll values the robot would use in practice. We conduct the calibration procedure outlined above independently for each arm at each angle, giving us two calibration tables for each arm. Thus, for a given arm and roll angle, we interpolate between these tables to generate a new table, which can then be used to conduct the bilinear interpolation method described above for any point on the board. After calibration, both arms of the robot reached positions on the checkerboard's plane with \SIrange{1}{2}{\milli\meter} of error for the two roll values tested.

\newcommand{\pegblocksubfloat}[2]{%
\subfloat[#1]{
  \adjustimage{width=228pt,origin=c,angle=-1.25,Clip=36pt}{#2}}}
  
\begin{figure*}[t]
\center
\begin{tabular}{c@{\quad}c@{\quad}c}
\pegblocksubfloat{RGB image}{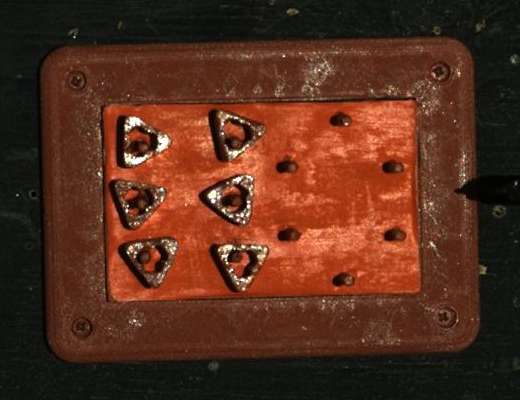} &
\pegblocksubfloat{Depth image}{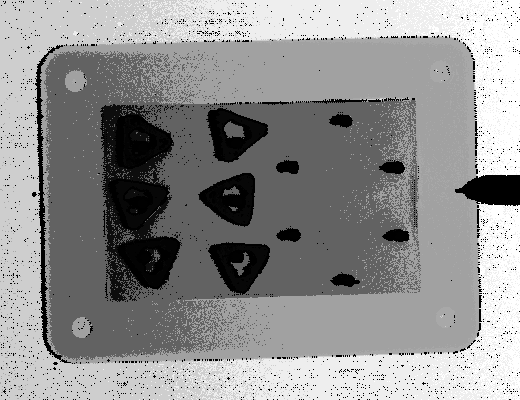} &
\pegblocksubfloat{Output of depth-sensing algorithm}{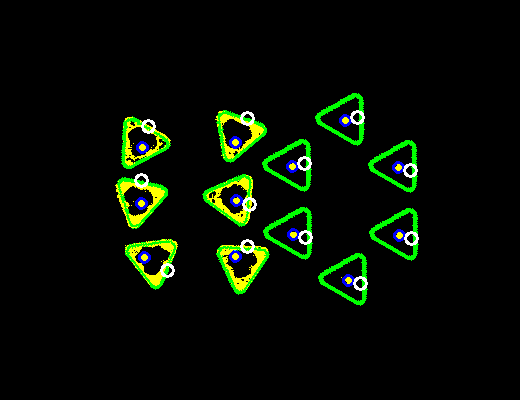}
\end{tabular}
\caption{
\small
Images as seen through the overhead RGBD camera (i.e., the depth sensor is not attached to the robot arm), and subsequently processed for later usage. Images (a) and (b) show sample RGB and depth images, respectively, that we might get at the beginning of an episode, where all pegs are informally dropped in the left half of the peg board. In (a), several blocks exhibit specular reflections that one might find in a surgical setting and could hinder other sensing methods. The board, pegs, and blocks are painted red. Image (c) shows the 12 detected pegs, circled in blue, along 6 grasp points (left half) and 6 place points (right half), all 12 of which are circled white. To the left, image (c) shows the location of the blocks that are within a pre-determined depth interval for block heights. Finally, 12 contours of the blocks are overlaid. To perform an episode, we follow a pre-determined order of grasp and place movements based on the white circles, and repeat the process going from right to left.
}
\vspace*{-6pt}
\label{fig:block-grasping}
\end{figure*}

\section{Depth-Sensing Algorithm}\label{sec:sensing}

\begin{algorithm}[htb!]
  \caption{Depth Sensing}
  \label{alg:sensing}
\begin{algorithmic}[1]
  \Statex{\textbf{Require:} Depth Image $I$, number of blocks $n$, block masks $\{M_{\phi_i}\}_{i=1}^k$, depth target $d$, depth tolerance $\epsilon$}
  \State $\mathtt{I}_\mathrm{thresh} = \mathtt{clip}(I, d-\epsilon, d+\epsilon) > 0$ // threshold depth
  \State // Get activation maps to find objects matching the masks
  \State $\{A_{\phi_i}\}_{i=1}^k = \{\mathtt{cross\_correlate(I_\mathrm{thresh}, M_{\phi_i})}\}_{i=1}^k$
  \For{$j \in \{1,\dots,n\}$}
    \State $\theta_j \leftarrow$ orientation $\phi_i$ of map with highest activation
    \State $p_j \leftarrow \arg\max_p A_{\theta_j}(p)$
    \State Zero out all activations in area of size of $M_{\theta_j}$ at $p_j$
    \EndFor
    \State Return $\{(p_j, \theta_j)\}_{j=1}^n$
\end{algorithmic}
\end{algorithm}

\begin{algorithm}[htb!]
  \caption{Grasp Planner}
  \label{alg:grasp_plan}
\begin{algorithmic}[1]
  \Statex{\textbf{Require:} Block pose $(p_\mathrm{block}, \theta)$, Arm $A \in \{\mathrm{left}, \mathrm{right}\}$, Peg location $p_\mathrm{peg}$}
  \State $(s_1, s_2) \leftarrow$ closest two (out of three) sides to $A$
  \State $\mathcal{G} \leftarrow 2$ grasp candidates along $s_1$ and $s_2$
  \State Return $\arg\max_{g\in \mathcal{G}} \|g - p_\mathrm{peg}\|_2$ 
\end{algorithmic}
\end{algorithm}

As shown in Fig.~\ref{fig:block-grasping} and Algorithm \ref{alg:sensing}, the block-detection algorithm uses a depth image from the overhead RGBD camera and thresholds it to only include the top surfaces of $n$ blocks by using a target depth value $d$ and a tolerance parameter $\epsilon$. In this work, $n=6$ but the algorithm can scale to larger values of $n$ as long as all blocks are reachable. The algorithm then cross-correlates the thresholded image with pre-computed masks $\{M_{\phi_i}\}_{i=1}^k$ of the block rotated in 30 different orientations in the plane of the image. The algorithm then proceeds iteratively to find the blocks. The best match at iteration $j$ is saved as a pixel coordinate $p_j = (u_j, v_j)$ and orientation $\theta_j$. The procedure zeroes out the region of the thresholded image occupied by the best match's mask then proceeds to the next iteration to find the next best match. We find the peg locations in a similar way. By computing cross-correlations with masks of the target objects (blocks, pegs), we are searching the image for objects that match the geometry of the target. Algorithm~\ref{alg:sensing} is implemented using SciPy~\cite{bressert2012scipy} signal processing code.

After detecting the blocks, we compute grasps (Algorithm~\ref{alg:grasp_plan}) by first sub-dividing each block edge into two potential grasp points, for a total of six potential grasp points per block.  We select the two sides that are closest to the end effector's current position; this is to prevent having the robot reach ``behind'' a peg. Of the grasp points on those edges, we select the grasp point furthest from the peg to get the most clearance for an open gripper and thus decreases the chances of collisions with pegs. Fig.~\ref{fig:block-grasping} shows an example setup of RGB and depth images and the corresponding proposed pick and place points circled in white, for the single-arm case.

\section{Trajectory Motion for Grasping}\label{sec:trajectory-motion}

\begin{figure*}[t]
\center
\includegraphics[width=1.00\textwidth]{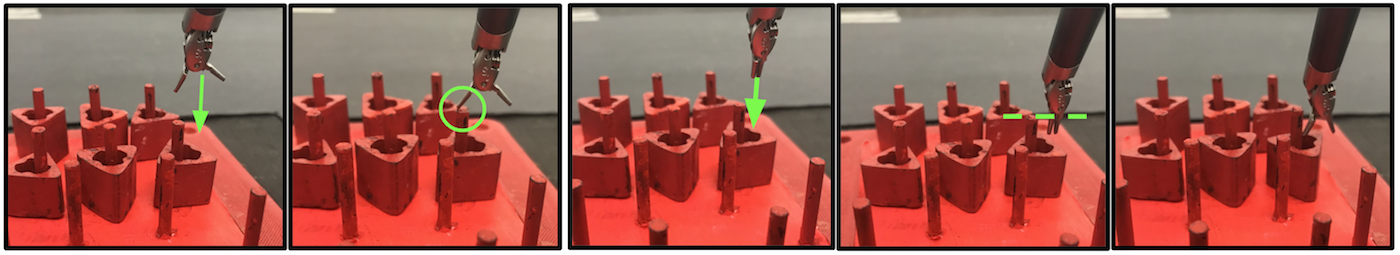}
\caption{
\small
A visualization of the dVRK's gripper as it picks up a block. In the first two images, an arm that descends to grasp a block with an open gripper can risk colliding with the top of the peg (circled above) and damage the hardware. To avoid this, as shown in the sequence in the last three images, the gripper is kept closed until the tip surpasses the height of the peg (dashed line) and then it opens, allowing for safer grasping of a block.
}
\vspace*{-10pt}
\label{fig:trajectory-motion}
\end{figure*}

%
%

After the block detection and grasp analysis, the system moves blocks from one set of pegs to another in a predefined order, iterating through motions from block grasp points to placement points until all blocks are moved.

For a given grasp point, the system commands the gripper to go slightly above the grasp point with a closed gripper. Then, the system opens the gripper, lowers it, closes the gripper, and raises it (ideally with the block). The system then moves the block over its placement point and opens the gripper, dropping the block on to a peg.   This full motion sequence is executed in an open loop without feedback or visual servoing, and thus depends on accurate calibration.

Extending to the bilateral-arm case, we select an ordering of the pegs/blocks for both arms such that they do not collide with each other during the trajectory; the left and right arms go to two neighboring pegs, such that each arm grasps the block closest to it.  By simultaneously commanding both arms motions, the robot never makes motions that cross over each other, thus avoiding arm-to-arm collisions.

The gripper remains closed during most motions, reducing the chances that motion, block detection, or calibration errors will result in the gripper colliding with a block or peg.
Initial peg transfer trials showed a danger in that an open gripper could hit the top of a peg, and the resulting force applied from the dVRK's arm to its gripper could damage a cable (Fig.~\ref{fig:trajectory-motion}).
The safety measure of keeping the gripper closed was sufficient to avoid gripper-on-peg collisions.

All motions in the peg transfer system require adding a bit of extra clearance to add a safety margin to overcome a limitation we encountered when commanding the robot's motions in terms of the gripper pose.  The robot internal control software translates poses into joint angles through the calculation of the inverse kinematics.  We have observed that, while this software achieves poses accurately, pose-to-pose motions are linear interpolations in joint angles, and not in the gripper's pose.  Due to the non-linear translation between pose and joint angles, this linear-in-joint-space interpolation results in non-linear motions of the end effector.  By adding extra clearance to every pose, we add a safety margin and reduce failures.  However, this may result in motions that are less efficient than possible were we to instead command the robot using more advanced motion planning and trajectory optimization techniques, as we plan to do in future work.


\subsection{Error Recovery Stage}\label{ssec:error_recovery}


When a pick failure occurs, the block often still remains in its original peg. After the first set of block transfer attempts, we perform a scan through the six known starting block positions, and detect if any still remain, and allow the dVRK a second attempt to grasp each block if needed. (In principle, this process could repeat ad infinitum, but we found that the robot would often make similar errors in subsequent actions, so we limit to two attempts.)

\section{Experiments and Results}\label{sec:results}

We initialize a peg transfer task episode by randomly dropping six blocks on the six pegs on the left side of the peg board, producing variation in the pose of each block. We evaluate failures as described in Section~\ref{ssec:failure-cases}, abbreviating failure modes as Pick (pick failures), Stuck (place-stuck failures) and Fall (place-fall failures). 


\subsection{Single-Arm Results}

\begin{table}[t]
\caption{
\small
Results from experiments for both the single and bilateral arm cases, with 20 and 5 episodes, respectively. We report the success rate over the number of total block transfer attempts (236 and 59, respectively), along with the average time for each of those attempts. In addition, we report the fraction of the three failure modes on each block transfer attempt.
}
\centering
\begin{tabular}{l | l r l l l}
Task & Success Rate & Time (s) & Pick & Stuck & Fall \\ \hline 
Single   & 0.869 (205/236) & 10.02 & 0.013 & 0.072 & 0.046 \\
Bilateral & 0.780 (46/59)  &  5.72 & 0.034 & 0.068 & 0.119 \\
\end{tabular}
\vspace*{-10pt}
\label{tab:dvrk-results}
\end{table}

We perform 20 episodes of the single arm case with the dVRK, and report results in Table~\ref{tab:dvrk-results}. 
Across all 20 episodes, the robot performed 236 total attempts at moving a block from one half of the board to another, each of which took 10.02 seconds on average. A ``perfect'' episode with no failures involves 12 total attempts (6 per direction). The number of attempts in a given episode may be higher or lower depending on if repeated grasp attempts are needed or if placing failures occur during the first set of six block transfer attempts. For those 236 block attempts, we recorded 31 failure cases, of which 3 were pick failures, 17 were place-stuck failures, and 11 were place-fall failures. The success rate for a single block attempt is thus $205 / 236$ = \SI{86.9}{\percent}. For 4 of the 20 episodes, the dVRK executed the entire task without failures. Each full episode lasted $118.2 \pm 9.4$ seconds, which is nearly twice as long compared to the human operator (Section~\ref{ssec:danyal}), and is in part due to the safety checks that are built into the motion planning. The project website contains videos of full episodes.

\subsection{Bilateral Results}

We also study a bilateral case, where the second arm is the same instrument type as the arm used in the single experiment, and run for five episodes. As the two arms can both move their blocks simultaneously, the average length of each full episode is shorter, and was timed at $67.6 \pm 7.3$ seconds. The bilateral case raises the possibility of having a failure case with collisions among the arms, but we did not experience any due to carefully chosen block orderings as described in Section~\ref{sec:trajectory-motion}.

Over 5 episodes, the success rate of block transfer attempts was $46 / 59$ = \SI{78.0}{\percent} (Table~\ref{tab:dvrk-results}). We ran these episodes after the single-arm case, and there may have been extra wear and tear. Furthermore, the bilateral results require slight adjustments of the placing angle for each block to avoid arms and blocks colliding with each other, which may increase the chances of placing errors. Across the 59 attempts, there were 2 pick, 4 place-stuck, and 7 place-fall failures.

\subsection{Human Surgeon Teleop}\label{ssec:danyal}

\begin{table}[t]
\caption{
\small
Physical experiments from Dr. Danyal Fer for the single and bilateral arm cases, with two episodes each. Results are presented in a similar manner as in Table~\ref{tab:dvrk-results}.
}
\centering
\begin{tabular}{l | l r l l l }
Task & Success Rate & Time (s) & Pick & Stuck & Fall \\ \hline 
Single    & 0.958 (23/24) & 5.08 & 0.000 & 0.042 & 0.000 \\
Bilateral & 0.833 (20/24) & 3.45 & 0.046 & 0.125 & 0.000 \\
\end{tabular}
\vspace*{-10pt}
\label{tab:danyal-1}
\end{table}

Dr. Danyal Fer, a surgical resident, performed two episodes of the single- and double-arm peg transfer tasks following the same experiment protocol and setup as the automated system. Table~\ref{tab:danyal-1} summarizes peg transfer results from Dr. Fer. In addition, Appendix~\ref{sec:danyal-nonpainted} has results on Dr. Fer's corresponding episodes for the task with standard, off-the-shelf FLS peg transfer materials that were not painted red, and thus were less sticky and allowed for more color cues.

Dr. Fer did not experience any place-fall failures in his episodes, but had one failure (place-stuck) in the single-arm case and four failures (one grasp, three place-stuck) in the bilateral case. Some of Dr. Fer's placing attempts resulted in the block hitting part of the target peg, but he recovered by raising the block and repeating the placing motion, thus avoiding failures. Dr. Fer completed the single- and bilateral-arm trials in $61.0 \pm 3.0$ and $41.5 \pm 4.9$ seconds, which is significantly faster than the automated system.

\subsection{Failure Cases}

The vast majority of the dVRK's failures were placing failures (28 for single-arm, 11 for double-arm), such as those shown in Figs.~\ref{fig:failure-place-stuck} and~\ref{fig:failure-place-fall}). Placing is challenging because, even if we command the robot to drop at a fixed target for each peg, the orientation of the gripped block varies at the time of release. Dr. Fer was able to more reliably avoid placing failures because he could react in real time in case his initial placing did not fully insert the block into a peg.

A place-stuck failure is not as severe as a fallen block, because a gentle nudge can slide the block in place. In the dVRK experiments, we observed that several place-stuck cases were unintentionally ``corrected'' by a subsequent action, which either knocked the block into the peg or removed an underlying block that prevented full insertion. If we count those ``corrected'' blocks as successes, the dVRK's success rate for the single- and bilateral-arm setups would be $210/236$ = \SI{89.0}{\percent} and $49/59$ = \SI{83.1}{\percent}.

\section{Discussion and Future Work}\label{sec:conclusions}

In this paper, we explore the potential for depth sensing in automating the FLS peg transfer task. We demonstrate a proof of concept of the procedure and show how using a high-quality depth camera and calibration can allow a da Vinci surgical robot to autonomously perform the task with \SI{86.9}{\percent} and \SI{78.0}{\percent} success rates (single- and bilateral-arms, respectively).
Results suggest depth-sensing can be effective for automated peg transfer but there remains a significant gap between automated and expert human performance.

In future work, we will address placing failures. We will use more sophisticated calibration~\cite{seita_icra_2018}, motion-planning~\cite{Latombe:1991:RMP:532147}, visual servoing~\cite{kragic_servoing_2002}, and error recovery techniques using the depth-sensor and tactile feedback from joint motor currents to enhance error detection precision and speed.  We will also explore the use of a surgical robot simulator to run reinforcement learning~\cite{gym,opensource_rl_surgical_2019,Sutton_2018} to train closed-loop controllers for the task, and will additionally use ideas from deep reinforcement learning for placing and insertion tasks~\cite{deepmind_insertion_2018,nair_insertion_2019}.

\section*{Acknowledgments}
{\small
This research was performed at the AUTOLAB at UC Berkeley in affiliation with the Berkeley AI Research (BAIR) Lab, Berkeley Deep Drive (BDD), the Real-Time Intelligent Secure Execution (RISE) Lab, the CITRIS ``People and Robots'' (CPAR) Initiative, and with UC Berkeley's Center for Automation and Learning for Medical Robotics (Cal-MR). The authors were supported in part by donations from SRI International, Siemens, Google, Toyota Research Institute, Honda, Intel, and Intuitive Surgical. The da Vinci Research Kit was supported by the National Science Foundation, via the National Robotics Initiative (NRI), as part of the collaborative research project ``Software Framework for Research in Semi-Autonomous Teleoperation'' between The Johns Hopkins University (IIS 1637789), Worcester Polytechnic Institute (IIS 1637759), and the University of Washington (IIS 1637444).  Daniel Seita is supported by a National Physical Science Consortium Fellowship. 
}


\appendices

\section{Human Operator on Non-Painted Case}\label{sec:danyal-nonpainted}

\begin{table}[h]
\caption{
\small
Physical experiments from Dr. Danyal Fer on the peg transfer task setup without the red paint. Results are presented in a similar manner as in Table~\ref{tab:dvrk-results}.
}
\centering
\begin{tabular}{l | l r l l l l}
Task & Success Rate & Time (s) & Pick & Stuck & Fall \\ \hline 
Single    & 1.000 (120/120) & 4.61 & 0.000 & 0.000 & 0.000 \\
Bilateral & 0.959 (117/122) & 2.71 & 0.016 & 0.025 & 0.000 \\
\end{tabular}
\vspace*{-10pt}
\label{tab:danyal-raw}
\end{table}

As an extra comparison, Table~\ref{tab:danyal-raw} reports results from Dr. Fer using the non-painted setup. This version allows for stronger color cues, avoids sticky red paint, and is closer to the standard FLS peg transfer task. Results indicate that Dr. Fer's performance in this case is superior to that of the painted setup, with success rates of \SI{100.0}{\percent} and \SI{95.9}{\percent} for the single and bilateral arm setups (versus \SI{95.8}{\percent} and \SI{83.3}{\percent} with red paint), along with faster block attempts of \SI{4.61}{\second} and \SI{2.71}{\second} (versus \SI{5.08}{\second} and \SI{3.45}{\second} with red paint). He completed episodes in $55.3 \pm 4.5$ and $33.0 \pm 3.4$ seconds, respectively.

\bibliographystyle{IEEEtranS}
\bibliography{example}

\begin{thebibliography}{10}
\providecommand{\url}[1]{#1}
\csname url@samestyle\endcsname
\providecommand{\newblock}{\relax}
\providecommand{\bibinfo}[2]{#2}
\providecommand{\BIBentrySTDinterwordspacing}{\spaceskip=0pt\relax}
\providecommand{\BIBentryALTinterwordstretchfactor}{4}
\providecommand{\BIBentryALTinterwordspacing}{\spaceskip=\fontdimen2\font plus
\BIBentryALTinterwordstretchfactor\fontdimen3\font minus
  \fontdimen4\font\relax}
\providecommand{\BIBforeignlanguage}[2]{{%
\expandafter\ifx\csname l@#1\endcsname\relax
\typeout{** WARNING: IEEEtranS.bst: No hyphenation pattern has been}%
\typeout{** loaded for the language `#1'. Using the pattern for}%
\typeout{** the default language instead.}%
\else
\language=\csname l@#1\endcsname
\fi
#2}}
\providecommand{\BIBdecl}{\relax}
\BIBdecl

\bibitem{haptic-feedback-surgery-2019}
A.~Abiri, J.~Pensa, A.~Tao, J.~Ma, Y.-Y. Juo, S.~J. Askari, J.~Bisley,
  J.~Rosen, E.~P. Dutson, and W.~S. Grundfest, ``{Multi-Modal Haptic Feedback
  for Grip Force Reduction in Robotic Surgery},'' \emph{Scientific Reports},
  vol.~9, no.~1, p. 5016, 2019.

\bibitem{compliance_2018}
F.~Alambeigi, Z.~Wang, R.~Hegeman, Y.-H. Liu, and M.~Armand, ``{A Robust
  Data-driven Approach for Online Learning and Manipulation of Unmodeled 3-D
  Heterogeneous Compliant Objects},'' in \emph{IEEE Robotics and Automation
  Letters (RA-L)}, 2018.

\bibitem{Ballantyne2003}
G.~Ballantyne and F.~Moll, ``{The da Vinci Telerobotic Surgical System: The
  Virtual Operative Field and Telepresence Surgery},'' \emph{Surgical Clinics
  of North America}, vol.~83, no.~6, pp. 1293--1304, 2003.

\bibitem{Beasley2012}
R.~A. Beasley, ``{Medical Robots: Current Systems and Research Directions},''
  \emph{Journal of Robotics}, 2012.

\bibitem{bressert2012scipy}
E.~Bressert, \emph{SciPy and NumPy: an overview for developers}.\hskip 1em plus
  0.5em minus 0.4em\relax " O'Reilly Media, Inc.", 2012.

\bibitem{gym}
G.~Brockman, V.~Cheung, L.~Pettersson, J.~Schneider, J.~Schulman, J.~Tang, and
  W.~Zaremba, ``{OpenAI Gym},'' 2016.

\bibitem{rating_peg_transfer_2017}
J.~D. Brown, C.~E. O'Brien, S.~C. Leung, K.~R. Dumon, D.~I. Lee, and
  K.~Kuchenbecker, ``{Using Contact Forces and Robot Arm Accelerations to
  Automatically Rate Surgeon Skill at Peg Transfer},'' in \emph{IEEE
  Transactions on Biomedical Engineering}, 2017.

\bibitem{danyal_fer_suggestion2}
S.~L. Chang, A.~S. Kibel, J.~D. Brooks, and B.~I. Chung, ``{The Impact of
  Robotic Surgery on the Surgical Management of Prostate Cancer in the USA},''
  \emph{BJU International}, vol. 115, no.~6, 2014.

\bibitem{improved_knots_case_2013}
D.-L. Chow and W.~Newman, ``{Improved Knot-Tying Methods for Autonomous Robot
  Surgery},'' in \emph{IEEE Conference on Automation Science and Engineering
  (CASE)}, 2013.

\bibitem{danyal_fer_suggestion}
B.~Chughtai, D.~Scherr, J.~D. Pizzo, M.~Herman, C.~Barbieri, J.~Mao, A.~Isaacs,
  R.~Lee, A.~E. Te, S.~A. Kaplan, P.~Schlegel, and A.~Sedrakyan, ``{National
  Trends and Cost of Minimally Invasive Surgery in Urology},'' \emph{Urology
  Practice}, vol.~2, no.~2, 2015.

\bibitem{fls_1998}
A.~M. Derossis, G.~M. Fried, M.~Abrahamowicz, H.~H. Sigman, J.~S. Barkun, and
  J.~L. Meakins, ``{Development of a Model for Training and Evaluation of
  Laparoscopic Skills},'' \emph{The American Journal of Surgery}, vol. 175,
  no.~6, 1998.

\bibitem{automated_needle_pickup_2018}
C.~D’Ettorre, G.~Dwyer, X.~Du, F.~Chadebecq, F.~Vasconcelos, E.~De~Momi, and
  D.~Stoyanov, ``{Automated Pick-up of Suturing Needles for Robotic Surgical
  Assistance},'' in \emph{IEEE International Conference on Robotics and
  Automation (ICRA)}, 2018.

\bibitem{fls_2004}
G.~M. Fried, L.~S. Feldman, M.~C. Vassiliou, S.~A. Fraser, D.~Stanbridge,
  G.~Ghitulescu, and C.~G. Andrew, ``{Proving the Value of Simulation in
  Laparoscopic Surgery},'' \emph{Annals of Surgery}, vol. 240, no.~3, 2004.

\bibitem{garg2016gpas}
A.~Garg, S.~Sen, R.~Kapadia, Y.~Jen, S.~McKinley, L.~Miller, and K.~Goldberg,
  ``{Tumor Localization using Automated Palpation with Gaussian Process
  Adaptive Sampling},'' in \emph{IEEE Conference on Automation Science and
  Engineering (CASE)}, 2016.

\bibitem{kalman_filter_2016}
M.~Haghighipanah, M.~Miyasaka, Y.~Li, and B.~Hannaford, ``{Unscented Kalman
  Filter and 3D Vision to Improve Cable Driven Surgical Robot Joint Angle
  Estimation},'' in \emph{IEEE International Conference on Robotics and
  Automation (ICRA)}, 2016.

\bibitem{raven2013}
B.~Hannaford, J.~Rosen, D.~Friedman, H.~King, P.~Roan, L.~Cheng, D.~Glozman,
  J.~Ma, S.~Kosari, and L.~White, ``{Raven-II: An Open Platform for Surgical
  Robotics Research},'' in \emph{IEEE Transactions on Biomedical Engineering},
  2013.

\bibitem{surgical_camera_2018}
J.~J. Ji, S.~Krishnan, V.~Patel, D.~Fer, and K.~Goldberg, ``{Learning 2D
  Surgical Camera Motion From Demonstrations},'' in \emph{IEEE Conference on
  Automation Science and Engineering (CASE)}, 2018.

\bibitem{knot_tying_2002}
H.~Kang and J.~Wen, ``{Robotic Knot Tying in Minimally Invasive Surgeries},''
  in \emph{IEEE/RSJ International Conference on Intelligent Robots and Systems
  (IROS)}, 2002.

\bibitem{surgical_robotics_2016}
Y.~Kassahun, B.~Yu, A.~T. Tibebu, D.~Stoyanov, S.~Giannarou, J.~H. Metzen, and
  E.~V. Poorten, ``{Surgical Robotics Beyond Enhanced Dexterity
  Instrumentation: a Survey of Machine Learning Techniques and Their Role in
  Intelligent and Autonomous Surgical Actions},'' \emph{Journal of Computer
  Assisted Radiology and Surgery}, 2016.

\bibitem{dvrk2014}
P.~Kazanzides, Z.~Chen, A.~Deguet, G.~Fischer, R.~Taylor, and S.~DiMaio, ``{An
  Open-Source Research Kit for the da Vinci Surgical System},'' in \emph{IEEE
  International Conference on Robotics and Automation (ICRA)}, 2014.

\bibitem{Kehoe2014}
B.~Kehoe, G.~Kahn, J.~Mahler, J.~Kim, A.~Lee, A.~Lee, K.~Nakagawa, S.~Patil,
  W.~Boyd, P.~Abbeel, and K.~Goldberg, ``{Autonomous Multilateral Debridement
  with the Raven Surgical Robot},'' in \emph{IEEE International Conference on
  Robotics and Automation (ICRA)}, 2014.

\bibitem{kragic_servoing_2002}
D.~Kragic and H.~I. Christensen, ``{Survey on Visual Servoing for
  Manipulation.}'' 2002.

\bibitem{krishnan2017ddco}
S.~Krishnan, R.~Fox, I.~Stoica, and K.~Goldberg, ``{DDCO: Discovery of Deep
  Continuous Options for Robot Learning from Demonstrations},'' in
  \emph{Conference on Robot Learning (CoRL)}, 2017.

\bibitem{Latombe:1991:RMP:532147}
J.-C. Latombe, \emph{Robot Motion Planning}.\hskip 1em plus 0.5em minus
  0.4em\relax Kluwer Academic Publishers, 1991.

\bibitem{super_yip_2019}
Y.~Li, F.~Richter, J.~Lu, E.~K. Funk, R.~K. Orosco, J.~Zhu, and M.~C. Yip,
  ``{SuPer: A Surgical Perception Framework for Endoscopic Tissue Manipulation
  with Surgical Robotics},'' \emph{arXiv:1909.05405}, 2019.

\bibitem{madapana2019desk}
N.~Madapana, M.~M. Rahman, N.~Sanchez-Tamayo, M.~V. Balakuntala, G.~Gonzalez,
  J.~P. Bindu, L.~Venkatesh, X.~Zhang, J.~B. Noguera, T.~Low \emph{et~al.},
  ``{DESK: A Robotic Activity Dataset for Dexterous Surgical Skills Transfer to
  Medical Robots},'' in \emph{IEEE/RSJ International Conference on Intelligent
  Robots and Systems (IROS)}, 2019.

\bibitem{mahler2014case}
J.~Mahler, S.~Krishnan, M.~Laskey, S.~Sen, A.~Murali, B.~Kehoe, S.~Patil,
  J.~Wang, M.~Franklin, P.~Abbeel, and K.~Goldberg, ``{Learning Accurate
  Kinematic Control of Cable-Driven Surgical Robots Using Data Cleaning and
  Gaussian Process Regression.}'' in \emph{IEEE Conference on Automation
  Science and Engineering (CASE)}, 2014.

\bibitem{mahler2017dexnet}
J.~Mahler, J.~Liang, S.~Niyaz, M.~Laskey, R.~Doan, X.~Liu, J.~A. Ojea, and
  K.~Goldberg, ``{Dex-Net 2.0: Deep Learning to Plan Robust Grasps with
  Synthetic Point Clouds and Analytic Grasp Metrics},'' in \emph{Robotics:
  Science and Systems (RSS)}, 2017.

\bibitem{mckinley2016}
S.~McKinley, A.~Garg, S.~Sen, D.~V. Gealy, J.~P. McKinley, Y.~Jen, M.~Guo,
  D.~Boyd, and K.~Goldberg, ``{An Interchangeable Surgical Instrument System
  with Application to Supervised Automation of Multilateral Tumor Resection},''
  in \emph{IEEE Conference on Automation Science and Engineering (CASE)}, 2016.

\bibitem{mckinley2015}
S.~McKinley, A.~Garg, S.~Sen, R.~Kapadia, A.~Murali, K.~Nichols, S.~Lim,
  S.~Patil, P.~Abbeel, A.~M. Okamura, and K.~Goldberg, ``{A Disposable Haptic
  Palpation Probe for Locating Subcutaneous Blood Vessels in Robot-Assisted
  Minimally Invasive Surgery},'' in \emph{IEEE Conference on Automation Science
  and Engineering (CASE)}, 2015.

\bibitem{miyasaka2015}
M.~Miyasaka, J.~Matheson, A.~Lewis, and B.~Hannaford, ``{Measurement of the
  Cable-Pulley Coulomb and Viscous Friction for a Cable-Driven Surgical Robotic
  System},'' in \emph{IEEE/RSJ International Conference on Intelligent Robots
  and Systems (IROS)}, 2015.

\bibitem{Moustris2011}
G.~Moustris, S.~Hiridis, K.~Deliparaschos, and K.~Konstantinidis, ``{Evolution
  of Autonomous and Semi-Autonomous Robotic Surgical Systems: A Review of the
  Literature},'' \emph{International Journal of Medical Robotics and Computer
  Assisted Surgery}, vol.~7, 2011.

\bibitem{tsc-dl-2016}
A.~Murali, A.~Garg, S.~Krishnan, F.~T. Pokorny, P.~Abbeel, T.~Darrell, and
  K.~Goldberg, ``{TSC-DL: Unsupervised Trajectory Segmentation of Multi-Modal
  Surgical Demonstrations with Deep Learning},'' in \emph{IEEE International
  Conference on Robotics and Automation (ICRA)}, 2016.

\bibitem{murali2015learning}
A.~Murali, S.~Sen, B.~Kehoe, A.~Garg, S.~McFarland, S.~Patil, W.~D. Boyd,
  S.~Lim, P.~Abbeel, and K.~Goldberg, ``{Learning by Observation for Surgical
  Subtasks: Multilateral Cutting of 3D Viscoelastic and 2D Orthotropic Tissue
  Phantoms},'' in \emph{IEEE International Conference on Robotics and
  Automation (ICRA)}, 2015.

\bibitem{Osa-RSS-14}
T.~Osa, N.~Sugita, and M.~Mitsuishi, ``{Online Trajectory Planning in Dynamic
  Environments for Surgical Task Automation},'' in \emph{Robotics: Science and
  Systems (RSS)}, 2014.

\bibitem{laparoscopic_transfer_2014}
L.~Panait, S.~Shetty, P.~Shewokits, and J.~A. Sanchez, ``{Do Laparoscopic
  Skills Transfer to Robotic Surgery?}'' \emph{Journal of Surgical Research},
  2014.

\bibitem{pastor2013}
P.~Pastor, M.~Kalakrishnan, J.~Binney, J.~Kelly, L.~Righetti, G.~Sukhatme, and
  S.~Schaal, ``{Learning Task Error Models for Manipulation},'' in \emph{IEEE
  International Conference on Robotics and Automation (ICRA)}, 2013.

\bibitem{intermittent_synch_2018}
V.~Patel, S.~Krishnan, A.~Goncalves, C.~Chen, W.~D. Boyd, and K.~Goldberg,
  ``{Using Intermittent Synchronization to Compensate for Rhythmic Body Motion
  During Autonomous Surgical Cutting and Debridement},'' in \emph{International
  Symposium on Medical Robotics (ISMR)}, 2018.

\bibitem{stewart_platform}
V.~Patel, S.~Krishnan, A.~Goncalves, and K.~Goldberg, ``{SPRK: A Low-cost
  Stewart Platform for Motion Study in Surgical Robotics},'' in
  \emph{International Symposium on Medical Robotics (ISMR)}, 2018.

\bibitem{rosen_icra_suturing_2017}
S.~A. Pedram, P.~Ferguson, J.~Ma, and E.~D.~J. Rosen, ``{Autonomous Suturing
  Via Surgical Robot: An Algorithm for Optimal Selection of Needle Diameter,
  Shape, and Path},'' in \emph{IEEE International Conference on Robotics and
  Automation (ICRA)}, 2017.

\bibitem{rosen_tissues_qlearn_2019}
S.~A. Pedram, P.~W. Ferguson, C.~Shin, A.~Mehta, E.~P. Dutson, F.~Alambeigi,
  and J.~Rosen, ``{Toward Synergic Learning for Autonomous Manipulation of
  Deformable Tissues via Surgical Robots: An Approximate Q-Learning
  Approach},'' in \emph{arXiv:1910.03398}, 2019.

\bibitem{rahmantransferring}
M.~M. Rahman, N.~Sanchez-Tamayo, G.~Gonzalez, M.~Agarwal, V.~Aggarwal, R.~M.
  Voyles, Y.~Xue, and J.~Wachs, ``{Transferring Dexterous Surgical Skill
  Knowledge between Robots for Semi-autonomous Teleoperation},'' in \emph{IEEE
  International Conference on Robot and Human Interactive Communication
  (Ro-Man)}, 2019.

\bibitem{opensource_rl_surgical_2019}
F.~Richter, R.~K. Orosco, and M.~C. Yip, ``{Open-Sourced Reinforcement Learning
  Environments for Surgical Robotics},'' in \emph{arXiv:1903.02090}, 2019.

\bibitem{ur5-peg-transfer-2019}
I.~Rivas-Blanco, C.~J.~P. del Pulgar, C.~López-Casado, E.~Bauzano, and V.~F.
  Munoz, ``{Transferring Know-How for an Autonomous Camera Robotic
  Assistant},'' \emph{Electronics: Cognitive Robotics and Control}, 2019.

\bibitem{auto_peg_transfer_2015}
J.~Rosen and J.~Ma, ``{Autonomous Operation in Surgical Robotics},''
  \emph{Mechanical Engineering}, vol. 137, no.~9, 2015.

\bibitem{saeidi_suturing_icra_2019}
H.~Saeidi, H.~N.~D. Le, J.~D. Opfermann, S.~Leonard, A.~Kim, M.~H. Hsieh, J.~U.
  Kang, and A.~Krieger, ``{Autonomous Laparoscopic Robotic Suturing with a
  Novel Actuated Suturing Tool and 3D Endoscope},'' in \emph{IEEE International
  Conference on Robotics and Automation (ICRA)}, 2019.

\bibitem{nair_insertion_2019}
G.~Schoettler, A.~Nair, J.~Luo, S.~Bahl, J.~A. Ojea, E.~Solowjow, and
  S.~Levine, ``{Deep Reinforcement Learning for Industrial Insertion Tasks with
  Visual Inputs and Natural Rewards},'' \emph{arXiv:1906.05841}, 2019.

\bibitem{Schulman2013}
J.~Schulman, A.~Gupta, S.~Venkatesan, M.~Tayson-Frederick, and P.~Abbeel, ``{A
  Case Study of Trajectory Transfer Through Non-Rigid Registration for a
  Simplified Suturing Scenario},'' in \emph{IEEE/RSJ International Conference
  on Intelligent Robots and Systems (IROS)}, 2013.

\bibitem{seita_fabrics_2019}
D.~Seita, A.~Ganapathi, R.~Hoque, M.~Hwang, E.~Cen, A.~K. Tanwani,
  A.~Balakrishna, B.~Thananjeyan, J.~Ichnowski, N.~Jamali, K.~Yamane, S.~Iba,
  J.~Canny, and K.~Goldberg, ``{Deep Imitation Learning of Sequential Fabric
  Smoothing Policies},'' in \emph{arXiv:1910.04854}, 2019.

\bibitem{seita_icra_2018}
D.~Seita, S.~Krishnan, R.~Fox, S.~McKinley, J.~Canny, and K.~Goldberg, ``{Fast
  and Reliable Autonomous Surgical Debridement with Cable-Driven Robots Using a
  Two-Phase Calibration Procedure},'' in \emph{IEEE International Conference on
  Robotics and Automation (ICRA)}, 2018.

\bibitem{sen2016automating}
S.~Sen, A.~Garg, D.~V. Gealy, S.~McKinley, Y.~Jen, and K.~Goldberg,
  ``{Automating Multiple-Throw Multilateral Surgical Suturing with a Mechanical
  Needle Guide and Sequential Convex Optimization},'' in \emph{IEEE
  International Conference on Robotics and Automation (ICRA)}, 2016.

\bibitem{rosen_icra_tissues_2019}
C.~Shin, P.~W. Ferguson, S.~A. Pedram, J.~Ma, E.~P. Dutson, and J.~Rosen,
  ``{Autonomous Tissue Manipulation via Surgical Robot Using Learning Based
  Model Predictive Control},'' in \emph{IEEE International Conference on
  Robotics and Automation (ICRA)}, 2019.

\bibitem{extraction_needles_2019}
P.~Sundaresan, B.~Thananjeyan, J.~Chiu, D.~Fer, and K.~Goldberg, ``{Automated
  Extraction of Surgical Needles from Tissue Phantoms},'' in \emph{IEEE
  Conference on Automation Science and Engineering (CASE)}, 2019.

\bibitem{Sutton_2018}
R.~S. Sutton and A.~G. Barto, \emph{{Introduction to Reinforcement Learning}},
  2nd~ed.\hskip 1em plus 0.5em minus 0.4em\relax Cambridge, MA, USA: MIT Press,
  2018.

\bibitem{thananjeyan2019safety}
B.~Thananjeyan, A.~Balakrishna, U.~Rosolia, F.~Li, R.~McAllister, J.~E.
  Gonzalez, S.~Levine, F.~Borrelli, and K.~Goldberg, ``{Safety Augmented Value
  Estimation from Demonstrations (SAVED): Safe Deep Model-Based RL for Sparse
  Cost Robotic Tasks},'' \emph{arXiv:1905.13402}, 2019.

\bibitem{thananjeyan2017multilateral}
B.~Thananjeyan, A.~Garg, S.~Krishnan, C.~Chen, L.~Miller, and K.~Goldberg,
  ``{Multilateral Surgical Pattern Cutting in 2D Orthotropic Gauze with Deep
  Reinforcement Learning Policies for Tensioning},'' in \emph{IEEE
  International Conference on Robotics and Automation (ICRA)}, 2017.

\bibitem{vandenBerg2010}
J.~{Van Den Berg}, S.~Miller, D.~Duckworth, H.~Hu, A.~Wan, X.~Fu, K.~Goldberg,
  and P.~Abbeel, ``{Superhuman Performance of Surgical Tasks by Robots using
  Iterative Learning from Human-Guided Demonstrations},'' in \emph{IEEE
  International Conference on Robotics and Automation (ICRA)}, 2010.

\bibitem{deepmind_insertion_2018}
M.~Vecerik, O.~Sushkov, D.~Barker, T.~Rothorl, T.~Hester, and J.~Scholz, ``{A
  Practical Approach to Insertion with Variable Socket Position Using Deep
  Reinforcement Learning},'' in \emph{IEEE International Conference on Robotics
  and Automation (ICRA)}, 2019.

\bibitem{yip2017robot}
M.~Yip and N.~Das, ``{Robot Autonomy for Surgery},'' \emph{The Encyclopedia of
  Medical Robotics}, 2017.

\bibitem{needle_insertion_deformation_2019}
F.~Zhong, Y.~Wang, Z.~Wang, and Y.-H. Liu, ``{Dual-Arm Robotic Needle Insertion
  With ActiveTissue Deformation for Autonomous Suturing},'' in \emph{IEEE
  Robotics and Automation Letters (RA-L)}, 2019.

\end{thebibliography}

\end{document}